# Exact Shapley Values for Local and Model-True Explanations of Decision Tree Ensembles

1. Thomas W. Campbell, Biodesix. Contributions: conceptualization, formal analysis, methodology, software, writing – original draft, writing – review and editing. Corresponding Author. Address: 2970 Wilderness Place, Suite 100, Boulder, CO 80301
2. Heinrich Roder, Biodesix. Contributions: conceptualization, writing – review and editing
3. Robert W. Georgantas III, Biodesix. Contributions: Generation of the project, supervision, writing – review and editing
4. Joanna Roder, Biodesix. Contributions: conceptualization, supervision, writing – review and editing

## Abstract

Additive feature explanations using Shapley values have become popular for providing transparency into the relative importance of each feature to an individual prediction of a machine learning model. While Shapley values provide a unique additive feature attribution in cooperative game theory, the Shapley values that can be generated for even a single machine learning model are far from unique, with theoretical and implementational decisions affecting the resulting attributions. Here, we consider the application of Shapley values for explaining decision tree ensembles and present a novel approach to Shapley value-based feature attribution that can be applied to random forests and boosted decision trees. This new method provides attributions that accurately reflect details of the model prediction algorithm for individual instances, while being computationally competitive with one of the most widely used current methods. We explain the theoretical differences between the standard and novel approaches and compare their performance using synthetic and real data.

## Acknowledgments

We thank Lelia Net and Laura Maguire for their useful discussion throughout this work and thoughtful review of the manuscript.

# Introduction

Decision tree ensembles remain among the most popular non-linear machine learning models in use today, particularly in medicine and biology [1] [2] [3] [4]. Decision trees have the appealing property that they can be directly interpreted – it is easy to understand how each feature or attribute is used in determining the prediction for a particular instance. However, ensemble structures preclude a straightforward direct interpretation of the importance of each feature to the overall model prediction. Shapley values (SVs) [5], an additive explanation assigning a relative importance to each feature which sum to the prediction for a particular instance, provide a framework to help interpret individual trees and tree ensembles. Since the introduction of TreeSHAP [6] [7], an algorithm allowing the computationally efficient generation of SVs for tree ensembles, it has become increasingly popular to include these additive explanations alongside standard performance estimates when evaluating model performance [8] [9] [10] [11] [12] [13].

In clinical applications, these explanations provide important information to guide physician-patient discussions when a model classifies a patient as high risk for some undesirable outcome [10] and to help understand biases learned by the model [14] [15] [16]. In finance, these explanations provide insight into why an individual loan applicant was denied or allowed credit when using automated credit risk assessment, allowing for more transparency in the process and identification of possible biases [17]. While explanations based on SVs have become increasingly popular, calculating them requires making both theoretical and implementational choices that have consequences for their practical utility.

In the language of cooperative games, SVs assess how much each player of a coalition of players contributes to an overall gain from the collective cooperation. The cooperative game and resulting SVs are defined by the set of $M$ players, $S$, and the *utility*, $u(s): 2^M \to \mathbb{R}$, which is the cooperative gain if only players in a coalition $s \subseteq S$ are present. Adopting this theory for explaining machine learning models, the players become the features in the data and the cooperative gain becomes the model's prediction on an instance. For the utility, there are not only players (features), but also possible values of the features. Formally, a different game is defined for each instance, but sometimes the utility is taken to additionally depend on the feature values. One needs to define this utility in terms of the machine learning model for an arbitrary subset of features in order to calculate the SVs.

The SHAP formalism [18] defines the utility for an arbitrary feature subset as a conditional expectation value of the model's prediction over the features not contained in the subset conditioned on the feature values of the instance. In practice, these expectation values are estimated observationally using some data set. This approach, which we will call *observational*, maintains the correlations between the features included in and excluded from the feature subset. It has been noted that observational approaches may produce some counterintuitive explanations [19] [20] [21]. In particular, the resulting SV explanations can assign non-zero importance to features not used at all in the model. In contrast, the *interventional* [6] [19] [20] formulation of SHAP breaks the dependence between the features included and excluded in the subset via Perl's *do* operation [22] in an attempt to account for correlations between features that might be present in the data itself rather than those learned by the model, even if the correlations are due to some latent variable the model does not see. While the interventional approach avoids the assignment of non-zero SVs to features not used in the model by breaking the feature correlational structure, the method involves off-manifold averaging, i.e., using sets of feature values that may never be observed in real instances, which may be undesirable [23] [24]. Other methods of defining utilities for feature subsets have also been suggested, such as retraining the model entirely on only the subset of features in each coalition [6] [10] [25].

Each of these approaches for feature subset utility definition affects the final SVs, so that, while SVs are a unique solution to an attribution problem in cooperative games under certain axioms [18], their application to the explanation of even a single machine learning model is far from unique, and one needs to consider carefully the consequences of the choice of utility on the resulting attributions. This multiplicity of possible SVs is explored in [20] [26] and additionally in [27], where the authors compare the observational conditional expectation values of SHAP with an interventional approach. They argue the interventional framework provides explanations that are more *true to the data,* and the explanations from the observational approach are more *true to the model.* Further, they suggest that model-true explanations are better suited for explaining the predictions of a particular trained model and data-true explanations are more appropriate for explaining patterns in the training data itself, subject to the model architecture.

In this paper we first explore considerations in the choice of the utility for decision trees while introducing a novel approach that is computationally competitive to TreeSHAP. We then compare our approach to TreeSHAP both theoretically and in synthetic and real data applications. In the spirit of [27], it is important to keep in mind the specific explanation problem one wishes to solve. Here, we restrict our attention to providing individualized, model-true explanations of decision trees, in the sense that we are primarily interested in instance level explanations of a specific trained decision tree or tree ensemble. This aim contrasts with learning more general patterns in the training data subject to some class of models, like those that can be elucidated by data-true explanation methods. Accordingly, we do not desire that our attributions completely explain inherent correlations or causal relationships present in the data itself (like in [28]); the attributions should only explain how the specific model uses the attributes available to it to make a prediction for a particular instance. We believe that the tree explanation method we present here will be of interest to those with similar goals, such as the individualized explanations for patients and loan applicants mentioned in the introduction.

## Shapley Utilities for Decision Trees

We propose a new definition of the utility for decision trees, where node values are first assigned to all internal nodes in the tree in the same fashion as was used for the terminal nodes during training. Then, while following the decision path for a specific instance, when a feature is encountered that is not present in the coalition, the value of the last node passed is returned. When the top feature in the tree is missing, the uninformative SV is returned. We will refer to this method as “Eject”. By “Ejecting” from the tree early when a feature that is not present in the coalition is encountered, rather than taking an expectation value as in interventional and observational TreeSHAP, this method avoids considering portions of the tree that would not have been used in predicting on the instance with the full feature set. While the interventional approach is now default in [29] when a reference set is provided, here we will refer to the TreeSHAP algorithm as presented in [7] as “TreeSHAP” and the interventional approach as “Interventional”.

Before comparing the approaches, we propose an additional SV property we believe to be desirable for local, model-true explanations. The dummy player axiom as presented in [20] states that if a feature is not used in a model, the attribution for that feature should be zero. For applications focused on local explanations, one might want to further restrict this idea such that if a feature is not used in prediction *for an individual instance*, the attribution for that instance for that feature should be zero. We will refer to this property as *local dummy player.* It can also be thought of as a restriction on the choice of utility such that any feature not used in classification for an instance should be a null player for that instance according to the null player axiom. It is worth noting that in the strict translation of the game theoretic utility, where there is a different game and set of utilities for each instance, the null player axiom guarantees the local dummy player. We will refer to features that are not used in prediction for an instance as locally null features for that instance.

This idea is motivated by directly interpreting the single tree in figure 1. For instances with $x_1 < t_1$, feature $x_3$ is not used in prediction, and accordingly we may want our attribution for $x_3$ to be zero. Imposing the local dummy player property guarantees that our SV explanations are consistent with this intuitive explanation. We will show here that neither TreeSHAP nor Interventional approaches satisfy this property.

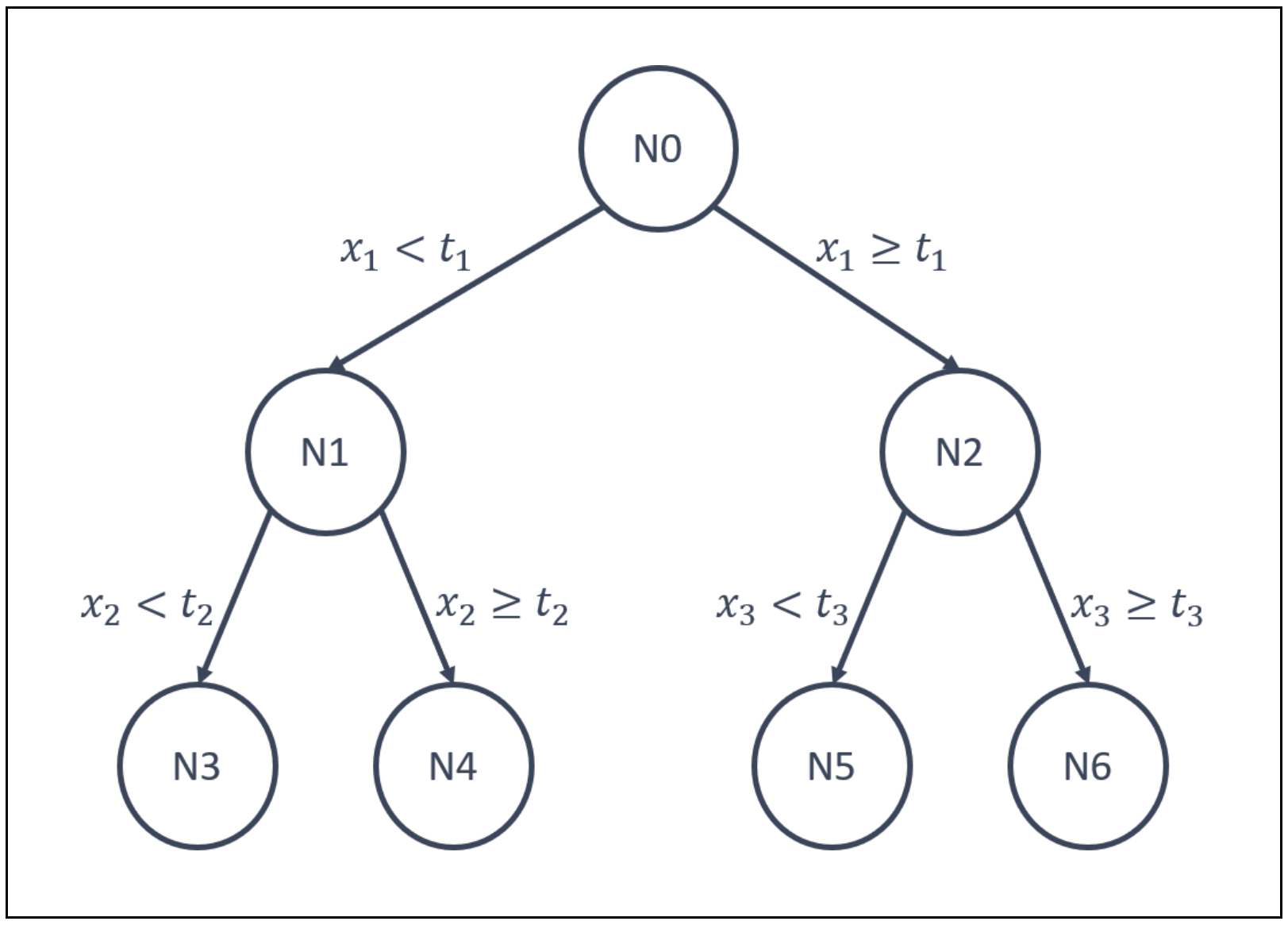

**Figure 1:** A simple decision tree with three features, $x_1$, $x_2$, and $x_3$ , with associated thresholds, $t_1$, $t_2$, and $t_3$ , and seven nodes, N0-N6.

In Figure 1, let $n_i$ be the number of training instances at node $N_i$, and suppose that $n_0$ is even with equal proportions of instances with a binary class label from {-1,1}. Suppose that each split results in proportions of the classes such that we assign the node values of -1 for nodes $N_1$, $N_3$, and $N_5$ and +1 for nodes $N_2$, $N_4$, and $N_6$. For an instance with $x_1 < t_1$, $x_2 < t_2$, and $x_3 \geq t_3$, table 1 gives the utilities and the SV of feature $x_3$ , $\phi_3$, for each approach.

| Coalition or SV | TreeSHAP | Interventional | Eject |
|---|---|---|---|
| $\{\}$ | $\frac{n_4 + n_6 - n_3 - n_5}{n_0}$ | $\frac{n_4 + n_6 - n_3 - n_5}{n_0}$ | $0$ |
| $\{x_1\}$ | $\frac{n_4 - n_3}{n_1}$ | $p(x_2 \geq t_2) - p(x_2 < t_2)$ | $-1$ |
| $\{x_2\}$ | $\frac{n_6 - n_5 - n_1}{n_0}$ | $\frac{n_6 - n_5 - n_1}{n_0}$ | $0$ |
| $\{x_3\}$ | $\frac{n_4 + n_2 - n_3}{n_0}$ | $\frac{n_4 + n_2 - n_3}{n_0}$ | $0$ |
| $\{x_1, x_2\}$ | $-1$ | $-1$ | $-1$ |
| $\{x_1, x_3\}$ | $\frac{n_4 - n_3}{n_1}$ | $p(x_2 \geq t_2) - p(x_2 < t_2)$ | $-1$ |
| $\{x_2, x_3\}$ | $\frac{n_2 - n_1}{n_0}$ | $\frac{n_2 - n_1}{n_0}$ | $0$ |
| $\{x_1, x_2, x_3\}$ | $-1$ | $-1$ | $-1$ |
| $\phi_3$ | $\frac{n_2 - n_6 + n_5}{2n_0}$ | $\frac{n_2 - n_6 + n_5}{2n_0}$ | $0$ |

**Table 1:** Utilities and SV for $x_3$ for an instance with $x_1 < t_1$, $x_2 < t_2$, and $x_3 \geq t_3$ for the three approaches, where $p(x_i < t_i)$ is the proportion of the interventional reference set (taken to be the training set for the tree) with $x_i < t_i$.

We notice that the TreeSHAP and Interventional approaches give similar, but not identical values for the utilities. We also see that, for this instance, $x_3$ is not a null player by TreeSHAP or Interventional as $u(\{x_2\}) \neq u(\{x_2, x_3\})$, while $x_3$ is a null player by Eject. Not surprisingly, $\phi_3 = 0$ for Eject, but is not necessarily 0 for TreeSHAP and Interventional.

As is perhaps apparent from the form of $\phi_3$ and how the calculation proceeds (formal algorithms in methods), the non-zero attribution from TreeSHAP and Interventional arises from off-decision path portions of the tree being included in their expectation values. Eject cannot do this by construction, guaranteeing that features off the decision path are null players. This property reduces the computational complexity of the SV evaluation (methods equation 1) for Eject, as the exponential sum over all feature subsets naturally reduces to an exponential sum over the set of unique features in the decision path. For a wide range of applications, such as boosted decision trees where the decision path often includes only a handful of features, this reduction can be quite significant.

Before presenting the results in detail, we provide a motivating example. Figure 2 compares SVs calculated using Eject and TreeSHAP for select features for an instance classified by an XGBoost [30] boosted decision tree trained to predict two-year disease-free survival on the breast cancer genomic data explored in a following section. The 10 features with largest mean magnitude SV across a validation cohort were selected for each approach and for each classification group. In interpreting these SVs, TreeSHAP suggests feature 3973 to be driving the prediction for this patient, while Eject assigns it very little attribution. Eject assigns high magnitude attribution to two features, 9572 and 2013, to which TreeSHAP assigns moderate magnitude attribution, but in the opposite direction. We notice that the instance level differences between the two approaches can be quite significant, potentially leading to different conclusions when interpreting the results.

The results we present here focus on comparing TreeSHAP with our Eject approach, but comparisons including the Interventional approach are given in the supplement. Generally, we observed that TreeSHAP and Interventional give nearly identical results, when compared to the differences we observe between TreeSHAP and Eject.

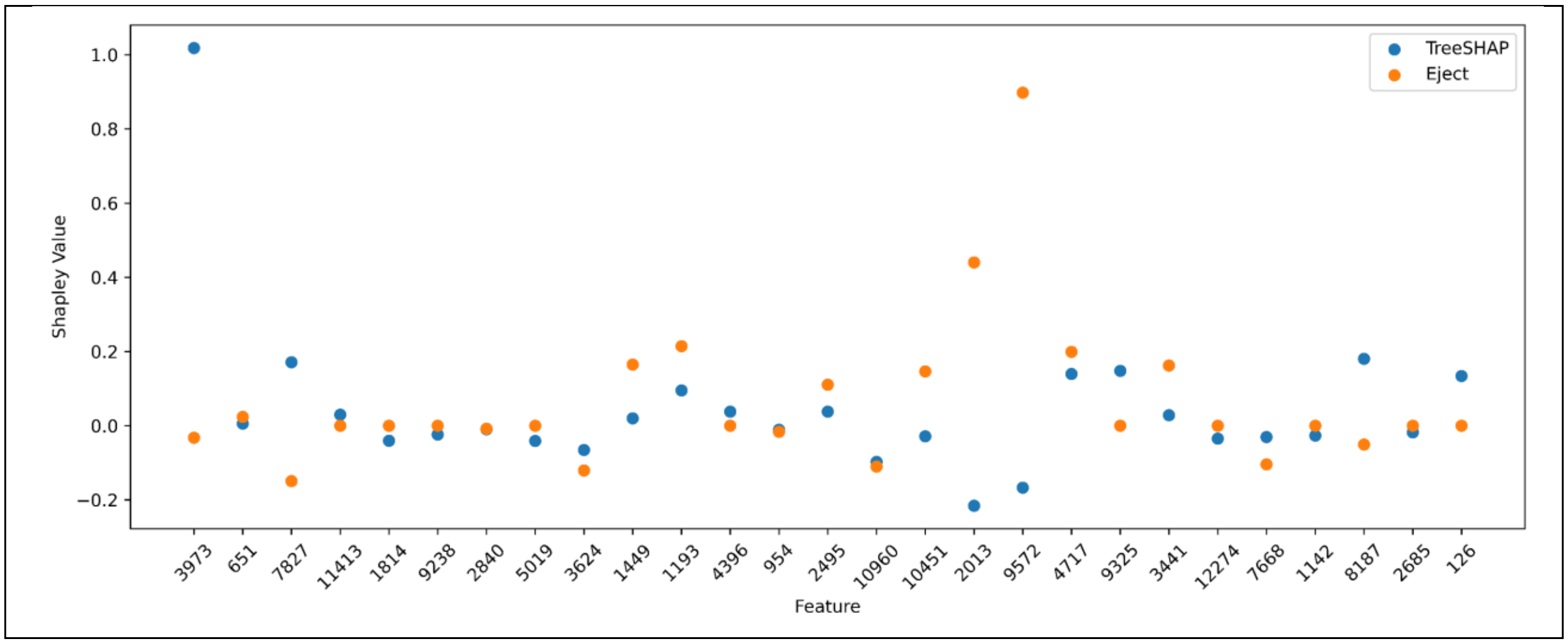

**Figure 2:** Eject and TreeSHAP SVs for selected features for an instance classified by an XGBoost boosted decision tree trained on breast cancer genomic data

## Synthetic Data Experiments

Synthetic data for binary classification problems were generated by drawing features from multivariate Gaussian distributions for two groups of instances, group 1 (training label of +1) and group 2 (training label of -1). For each feature, the means of the distributions for each group were either the same (uninformative features) or different (informative features) by a varying amount that we will refer to as expression difference. The variance of each feature was 1, and informative features could be correlated by a prespecified covariance matrix.

For each of 4 expression differences (0.25, 0.50, 0.75, and 1.0), a set of synthetic data with 200 uninformative features and 200 informative features (with equal expression difference) was generated, with 120 instances drawn randomly from each group. A random classification forest was trained on each set, and SVs were calculated on an independent set with 30 instances drawn randomly from each group. Accuracy in the validation set was 0.8 for the 0.25 expression difference set and 1.0 for the other three.

Figure 3 gives distributions of the product of the SVs and the instance classification ($\pm 1$) for both Eject and TreeSHAP. This product, denoted as SV~, is positive when an attribution is in the same direction as classification. Shaded areas represent a smoothed density distribution and the vertical lines give the quartiles.

As expected, TreeSHAP assigns non-zero attribution to locally null features, and these attributions can be large in magnitude for the informative features, while being smaller in magnitude for the uninformative ones. This observation suggests that TreeSHAP is providing an explanation more data-true and less model-true than Eject, where features that are not used in prediction are assigned zero attribution.

It is not unexpected for uninformative features to be used in the tree, particularly at the bottom, which is an important concept to keep in mind depending on the explanation problem. If we wish to provide more model-true explanations, we should be interested to know how much a seemingly uninformative feature contributed to prediction, whereas for more data-true applications, we may want our attributions to inform us that the uninformative feature generally contains little or no information useful for prediction. If an informative feature is used along the decision path and elsewhere in the tree or forest, it will likely have a similar effect in both places, whereas for uninformative features, on average over the multiple places in the forest where the feature is used, the effect should tend to cancel out. Because TreeSHAP considers off decision path portions of the tree, we suspect these effects may drive attribution higher for informative features and lower for uninformative ones, again providing a more data-true explanation compared to Eject, at the expense of not being model-true for a specific instance.

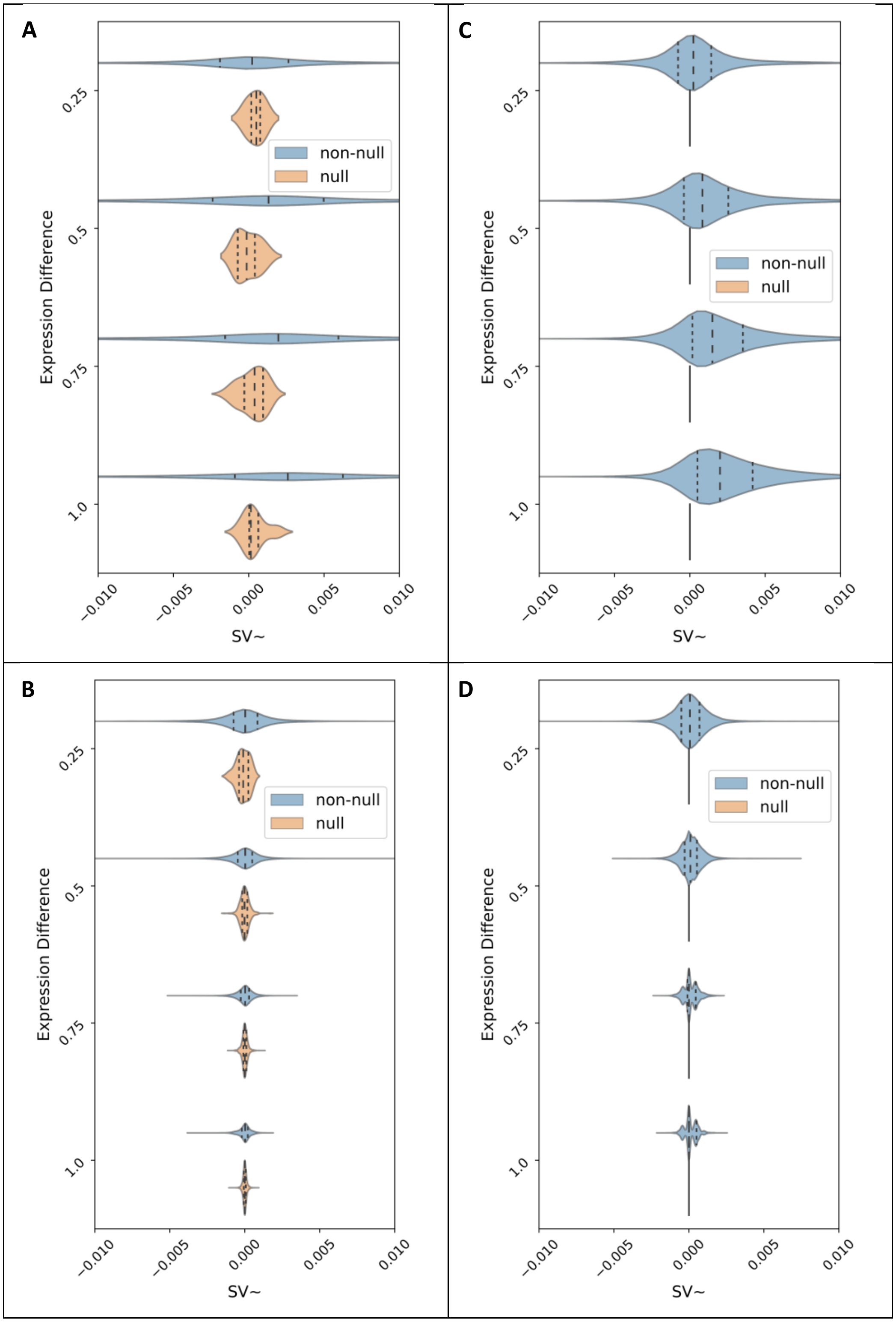

**Figure 3:** SV~ (SVs multiplied by the sign of instance prediction) distributions for the 4 experiments separated by locally null and non-null features. **A**: TreeSHAP, informative features. **B**: TreeSHAP, uninformative features. Vertical lines for null feature distributions indicate all values are exactly zero. **C**: Eject, informative features. **D**: Eject, uninformative features.

## Breast Cancer Genomics Experiment

The different approaches were compared in real-world data using two mRNA expression data sets for patients with non-metastatic breast cancer with 12,770 features and 295 patients in the training set and 380 in the validation set. A random classification forest with 10,000 trees was trained to predict two-year disease-free survival following surgery on the 295-patient set. The forest achieved 68% accuracy in the 380-patient independent validation set, and SVs were calculated for the validation set instances. For each predicted classification group and for both TreeSHAP and Eject, the five features with the largest mean absolute value SV were identified, and SVs for the union of those sets of features are given in figure 4A (TreeSHAP) and figure 4B (Eject) for the validation set. Generally, we observe that the TreeSHAP values have larger variance across instances for the selected features compared to Eject, and features 12234, 6481, and 8246 also show significant differences in their distribution.

For each instance and for each locally null feature, the TreeSHAP SV was normalized by the mean absolute SV for the instance and the distribution of these normalized SVs across instances and locally null features is shown in figure 4C. We see the TreeSHAP SVs for the locally null features can be comparable in magnitude to the mean magnitude over all features. We see this effect highlighted for a single instance. Figure 4D shows the 10 most positive and 10 most negative SVs for locally null features together with those for 20 locally non-null features chosen at random, along with the mean of all positive and negative values shown as the horizontal lines.

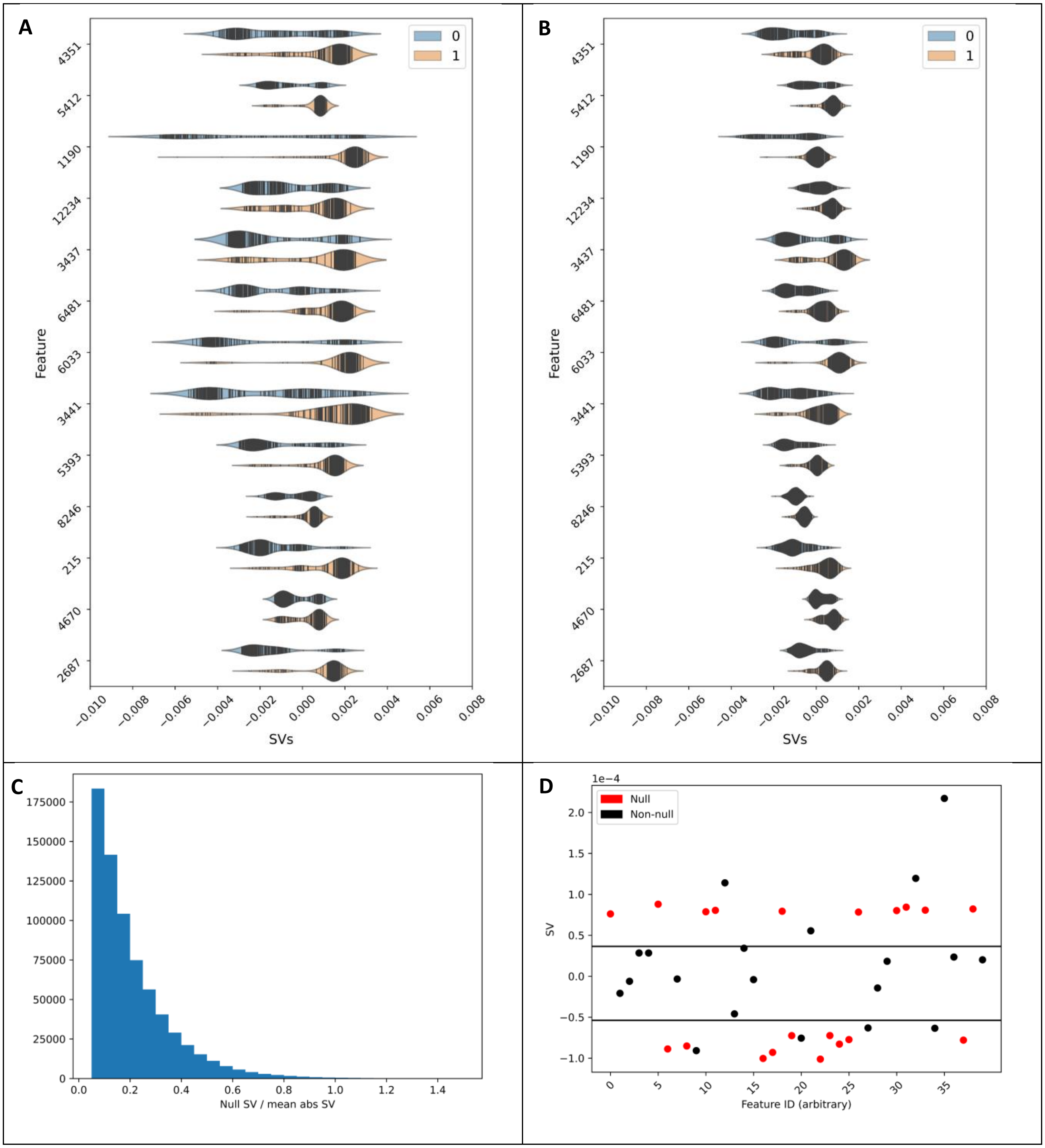

**Figure 4:** SVs for selected features for instances in the validation set using (**A**) TreeSHAP and (**B**) Eject. Shaded areas represent a smoothed density distribution of the SVs with the vertical lines being each individual SV. A class label of 1 indicates predicted disease progression. (**C**) the distribution of TreeSHAP SVs normalized by mean absolute value SV per instance for locally null features. (**D**) TreeSHAP SVs for a single instance for selected features with horizontal lines at the mean of all positive and negative SVs.

## NHANESI Mortality Experiments

The methods were compared on another real-world data set, the NHANESI I set [31], which was also used in [6] [7]. The features and mortality data were accessed from the SHAP repository [29]. Patients who had complete data for a selected set of 27 attributes (details in methods) were randomly assigned to a training cohort with 500 patients and a validation cohort of 659 patients. An XGBoost [30] boosted classification tree with 100 estimators and a maximum depth of five was trained on the training cohort to predict 10-year survival and achieved 83% accuracy in the validation cohort, where SVs for the different methods were also calculated. The log-loss objective function was used, so SVs are given in the log-loss space prior to transformation to a probability-like number by a sigmoid function. Together with the encoding in the data, this gives a negative value contribution to a classification of no death and a positive value contribution to a death prediction. For patients in the validation set predicted to die and for selected features, figure 5A-C gives heatmaps of TreeSHAP and Eject SVs along with their difference. Features were selected for inclusion in the plots by taking the union of the sets of the five features with the largest mean absolute SV for Eject and TreeSHAP. The differences between approaches at the instance level are clearly seen in the difference plot (Figure 5C). For example, for patients 0-5, TreeSHAP gives SVs for age roughly 50% larger than Eject. Figure 5D-F overlays SVs for both approaches for three patients shown in the heatmaps (5, 30, and 55). In Figure 5D, we see large differences in the SVs of all features shown, except for blood pressure and eosinophils, with the two approaches disagreeing on the largest magnitude feature. Figures 5G and 5H give the SVs for pulse pressure and lymphocytes, respectively, plotted against the value of the feature for each sample in the validation set. Here we notice that across a whole set, one may identify trends when looking at TreeSHAP SVs: higher pulse pressure being associated with lower mortality risk and low or high lymphocytes being associated with higher mortality risk, that are not so clear for the Eject SVs. Hence, one may draw different conclusions on the set and instance level, depending on the approach used. A SciKitLearn [32] random classification forest was also studied. Results were similar to those for the boosted classification tree and are given in the supplement.

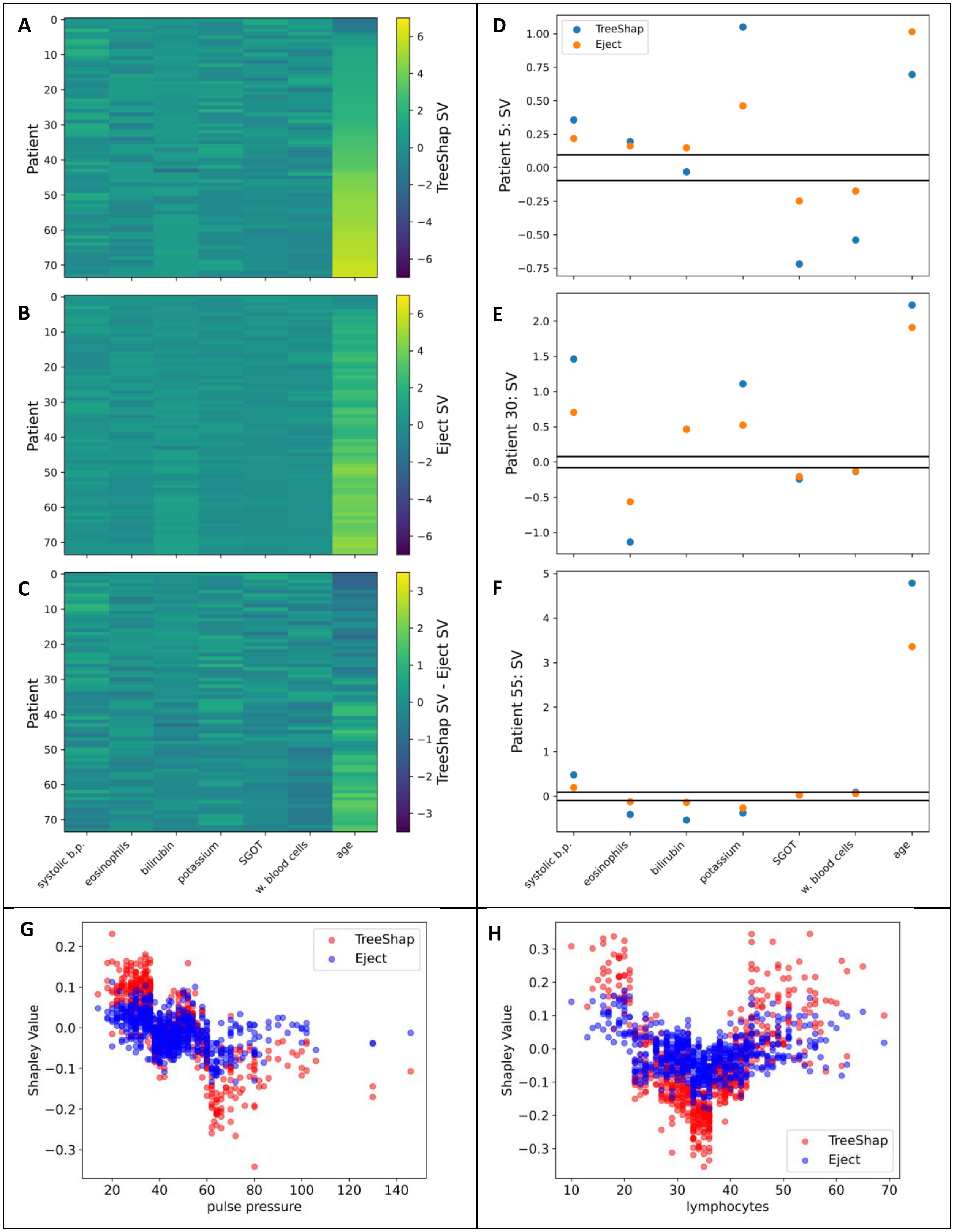

**Figure 5:** (**A**-**C**) heatmaps showing TreeSHAP and Eject SVs and their difference for selected features. (**D**-**F**) SV comparisons for patients: 5, 30, and 55. Horizontal lines given at $\pm 1/27$ (27 being the number of features used) for reference. (**G**-**H**) SV vs feature value for pulse pressure and lymphocytes.

## Discussion

With the growing popularity of explanations for tree ensemble methods, the choices in adopting SVs for these explanations should be carefully considered, as they may have conclusion-altering effects on the resulting attributions. We have proposed a new definition for the Shapley utility, the Eject approach, that provides highly model-true local explanations of decision trees by guaranteeing a local null player property that we believe is desirable in a wide range of explanation problems. Explanations of clinical decision support systems, where physicians may be highly interested in explanations for an individual patient to help guide treatment options, may benefit more from this or similar approaches than from TreeSHAP. Additionally, for credit risk systems, where unbiased and transparent explanations of the actual model making the decisions are a moral imperative, truly individualized and model-true explanations are a necessity.

We have compared this new approach with both the observational and interventional TreeSHAP approaches theoretically on a simple, directly interpretable, single tree and in ensemble models trained on synthetic and real-world data. We showed that both TreeSHAP approaches do not ensure a desirable local dummy player property. Non-zero attributions for features that are not used at all in generating the instance prediction may result, and these non-zero attributions may be large compared with the attributions of locally non-null features. For our experiments, we observed that observational and interventional TreeSHAP produce very similar attributions, while the Eject approach gives attributions that are different, both on the individual instance level and in aggregate across a cohort. These differences can potentially be large enough in magnitude to alter the conclusions drawn from interpreting the attributions, especially at the individual instance level. We hypothesized that these differences may be driven by both TreeSHAP approaches using off-prediction-path portions of the trees when assigning attribution. We believe our method to be superior for highly model-true explanations with a focus on individualized explanations, as it adheres to the local dummy player property.

An implementation of the Eject approach, given at [33], currently allows SV calculations using XGBoost boosted classification trees and sklearn random classification forests. The run times of Eject and TreeSHAP as implemented in [33] are compared in the methods, where we conclude that the Eject approach is computationally competitive for a wide range of applications, particularly for shallow trees that are common in boosted decision tree models.

When the details of a model and its architecture are not available, model-agnostic explainability approaches are invaluable. However, very often explanations are now presented as an integral part of a description of the development and validation of a model [8] [9] [10] [11] [12] [13]. Hence, the full model is available and consideration of the interplay of the approach taken to define utilities for SV calculations with model architecture is possible. In these cases, we strongly agree with authors [26] who have pointed out that SV calculations benefit from tailoring to the model architecture, as well as to the intended use of the attributions. We encourage others to carefully consider the consequences of choices made in adopting SVs for machine learning explainability.

# Methods

## Shapley Values

For a set, $S$, of $M$ players, the Shapley value for player $i$, $\phi_i$, is given by:

$$\phi_i = \sum_{s \subseteq S \setminus \{i\}} \frac{1}{M \binom{M-1}{|s|}} (u(s \cup \{i\}) - u(s)) \tag{1}$$

The *utility*, $u(s): 2^M \to \mathbb{R}$, is the cooperative gain if only players in the coalition $s$ are present. Shapley values are the unique additive solution to this attribution problem that satisfy four axioms, efficiency, symmetry, linearity, and null player, which are given below.

**Efficiency:**

$$\sum_i \phi_i = u(S)$$

**Symmetry:**

If for every $s \subseteq S \setminus \{i, j\}$, $u(s \cup \{i\}) = u(s \cup \{j\})$, then $\phi_i = \phi_j$

**Linearity:**

For utilities $u$, $v$, and real numbers $\alpha$, $\beta$:

$$\phi_i(\alpha * u + \beta * v) = \alpha * \phi_i(u) + \beta * \phi_i(v)$$

**Null Player:**

If for every $s \subseteq S \setminus \{i\}$, $u(s \cup \{i\}) = u(s)$, then $\phi_i = 0$

The game theoretic utility maps from a configuration of players to a real number; for explaining a machine learning model, there are not only players (features), but also possible values of those features for an instance. Adopting a similar notation as in [34], the utilities now depend on the feature values of instance, $x$ and feature set, $s$: $u(x, s)$, or, alternatively, one can consider that the values of the features for an instance define a distinct game for that instance.

## Software and Data

All software needed to reproduce the results presented here is available at [33], including python classes for training trees, evaluating SVs using the three approaches, and producing plots like those shown here. All datasets are also provided as used here, in addition to the script for the synthetic data generation. Experiments were run using python 3.8.5, included with all dependencies needed from Anaconda 4.9.2 [35].

For the synthetic data and breast cancer experiments, a simple CART implementation included in the github repository was used to train the forests. Continuous features were first binned by equal density into 10 bins for the search. The number of features searched at each node was floor(sqrt(total number of features)) and then a standard CART procedure optimized the cross-entropy at each node. Splits of the data were required to assign at least 5 instances to each daughter node for the split to be considered in the search. Class labels were assigned in the terminal nodes by majority vote over the training instances present.

For each tree in the forest, floor(2*N/3) instances were drawn without replacement from the N available in the training set and a tree was grown on that sampling. 10,000 trees were used in each forest to produce an ensemble prediction of the proportion of trees classifying the instance to the positive class. A threshold of 0.5 was used for producing classification labels, and the raw output was taken to be the simple average of the individual tree predictions, so that a simple average over the forest could also be used for SVs of the ensemble.

The XGBoost [30] boosted classification forest for the NHANESI experiment used the XGBClassifier class with max_depth=5 and n_estimators=100. The shap.explainers._tree.XGBTreeModelLoader class from the shap package was used to extract the necessary information from the individual trees for calculating SVs.

The sklearn [32] random classification forest for the NHANESI experiment given in the supplement used the RandomForestClassifier class with min_samples_leaf=5 and n_estimators=1000. The shap.explainers._tree.SingleTree class from the shap package was used to extract the necessary information from the individual trees for calculating SVs.

Matplotlib.pyplot [36] was used for figures 2, 4, and 5, and seaborn [37] was used for figures 3 and 4.

The breast cancer experiments used two sets of mRNA expression data generated from tissue samples collected from non-metastatic breast cancer patients at time of surgery. These data were also used in [38] and were collected from materials in [39]. The NKI set was used for training the model and consisted of 13,108 genetic features for 295 patients. The LOI set was used for model validation and the calculation of SVs, and it consisted of 17,585 genetic features for 380 patients. The sets had 12,770 common genomic features and each set was reduced to include only these features for use here. The ComBat tool [40] had been used to standardize the two sets from their original values to the version accessed from [38].

The NHANESI I data set was accessed through [29]. Only patients complete in the attributes listed in table 2 were included and all modelling used only those same attributes.

| | | |
|---|---|---|
| sex_isFemale | total_bilirubin | band_neutrophils |
| age | red_blood_cells | cholesterol |
| physical_activity | white_blood_cells | urine_pH |
| alkaline_phosphatase | hemoglobin | uric_acid |
| SGOT | hematocrit | systolic_blood_pressure |
| BUN | segmented_neutrophils | pulse_pressure |
| calcium | lymphocytes | bmi |
| creatinine | monocytes | |
| potassium | eosinophils | |
| sodium | basophils | |

**Table 2:** Selected attributes used from the NHANESI I data

The TreeSHAP implementation was adopted from the high-level version rather than the optimized implementation from [29]. The resulting SVs were adjusted so that $\phi_0 = 0$ , so that they were directly comparable to the Eject values. Interventional TreeSHAP was implemented from scratch using Algorithm 2 given in the next section.

## Utility Algorithms

Algorithms for each of the three utilities considered are given below so that their differences can be formally and transparently observed. Adopting the same notation as in [7], for a tree with node values in vector $v$, indices of the left and right child nodes in vectors $a$ and $b$ respectively, feature indices in vector $d$, thresholds in vector $t$, and the training set cover (proportion of instances in training flowing through a given node) in vector $r$, the TreeShap utility [6] for instance $x$ on subset $S$ is given in algorithm 1. Given a reference set $X$, the Interventional utility [6] is given in algorithm 2. We propose a third utility, Eject, given in algorithm 3.

**Algorithm 1** TreeSHAP

**procedure** $u_{\text{TS}}(x,\ S,\ \text{tree}=\{v,a,b,t,r,d\})$
  **procedure** $G_{\text{TS}}(j,\ w)$
    **if** $v_j \neq$ internal **then**
      **return** $w \cdot v_j$
    **else**
      **if** $d_j \in S$ **then**
        **if** $x_{d_j} \leq t_j$ **then**
          **return** $G_{\text{TS}}(a_j, w)$
        **else**
          **return** $G_{\text{TS}}(b_j, w)$
        **end if**
      **else**
        **return** $G_{\text{TS}}(a_j, \frac{w \cdot r_{a_j}}{r_j}) + G_{\text{TS}}(b_j, \frac{w \cdot r_{b_j}}{r_j})$
      **end if**
    **end if**
  **end procedure**
  **return** $G_{\text{TS}}(1,1)$
**end procedure**

**Algorithm 2** Interventional

**procedure** $u_{\text{INT}}(x,\ S,\ \text{tree}=\{v,a,b,t,r,d\},\ X)$
  **procedure** $G(j,\ x)$
    **if** $v_j \neq$ internal **then**
      **return** $v_j$
    **else**
      **if** $x_{d_j} \leq t_j$ **then**
        **return** $G(a_j)$
      **else**
        **return** $G(b_j)$
      **end if**
    **end if**
  **end procedure**
  $o \leftarrow 0$
  **for** $y \in X$ **do**
    $z \leftarrow x$
    **for** $i \leftarrow 1$ **to** $length(x)$ **do**
      **if** $i \notin S$ **then**
        $z_i \leftarrow y_i$
      **end if**
    **end for**
    $o \leftarrow o + G(1, z)$
  **end for**
  **return** $o/length(X)$
**end procedure**

**Algorithm 3** Eject

**procedure** $u_{\mathrm{EJ}}(x,\ S,\ \text{tree}=\{v,a,b,t,r,d\})$
    **procedure** $G_{\mathrm{EJ}}(j)$
        **if** $d_j \in S$ **then**
            **if** $x_{d_j} \leq t_j$ **then**
                **return** $G_{\mathrm{EJ}}(a_j)$
            **else**
                **return** $G_{\mathrm{EJ}}(b_j)$
            **end if**
        **else**
            **return** $v_j$
        **end if**
    **end procedure**
    **return** $G_{\mathrm{EJ}}(1)$
**end procedure**

## Computational Performance Comparison

For calculating SVs for a single instance on a single tree, the Eject algorithm is easily seen to be of exponential computational complexity in the number of unique features along the decision path, while the fast TreeSHAP algorithm given as algorithm 2 in [7] is quadratic in the tree depth, $D$, and linear in the number of leaves, $L$. Depending on the tree and the instance, either can be more complex, as the number of unique features along the decision path is less than or equal to the depth of the tree, giving formal upper bounds on complexity of $O(LD^2)$ for TreeSHAP compared to $O(2^D)$ for Eject. For shallow trees, such as those common in boosted decision tree ensembles, the two are clearly comparable.

To get a rough idea of how the two algorithms compare in practice, a simple comparison was performed using the Eject algorithm as implemented in [33] and the high-level python implementation of TreeSHAP also in [33], which was adopted from [29]. While this is certainly not a perfectly level comparison, both use similar level python libraries (i.e., numpy) and neither is highly optimized for performance like the actual implementation of TreeSHAP algorithm 2 in [29].

Using the NHANESI data, as employed throughout this study, a random classification forest with 100 trees was trained to predict mortality using the same parameters as the synthetic data experiments. SVs were then calculated using both algorithms for every feature on the entire validation cohort. The average CPU time for TreeSHAP was 8.4 times that for Eject. Using the breast cancer data, as employed throughout this study, and an identical forest to the previous comparison, the trend reversed, with the average CPU time for Eject being 5.6 times for that of TreeSHAP. We suspect this reversal could be due to the very large number of features causing repeat features in the decision path to be less likely, resulting in a larger unique set of features on the decision path. Using the synthetic data presented in the supplement and identical random forest parameters, the average CPU time for TreeSHAP was 10.7 times that for Eject.

# References

[1] R. Alsagheer, Ha, F. Abbas and A. S. A.-H. Alharan, "Popular decision tree algorithms of data mining techniques: a review," *International Journal of Computer Science and Mobile Computing,* vol. 6, p. 133–142, 2017.

[2] D. Che, Q. Liu, K. Rasheed and X. Tao, "Decision Tree and Ensemble Learning Algorithms with Their Applications in Bioinformatics," in *Advances in Experimental Medicine and Biology*, Springer New York, 2011, p. 191–199.

[3] A. Navada, A. N. Ansari, S. Patil and B. A. Sonkamble, "Overview of use of decision tree algorithms in machine learning," in *2011 IEEE Control and System Graduate Research Colloquium*, 2011.

[4] M. Somvanshi, P. Chavan, S. Tambade and S. V. Shinde, "A review of machine learning techniques using decision tree and support vector machine," in *2016 International Conference on Computing Communication Control and automation (ICCUBEA)*, 2016.

[5] L. S. Shapley, "17. A Value for n-Person Games," in *Contributions to the Theory of Games (AM-28), Volume II*, Princeton University Press, 1953, p. 307–318.

[6] S. M. Lundberg, G. Erion, H. Chen, A. DeGrave, J. M. Prutkin, B. Nair, R. Katz, J. Himmelfarb, N. Bansal and S.-I. Lee, "From local explanations to global understanding with explainable AI for trees," *Nature Machine Intelligence,* vol. 2, p. 56–67, January 2020.

[7] S. M. Lundberg, G. G. Erion and S.-I. Lee, "Consistent Individualized Feature Attribution for Tree Ensembles," February 2018.

[8] R. Agius, C. Brieghel, M. A. Andersen, A. T. Pearson, B. Ledergerber, A. Cozzi-Lepri, Y. Louzoun, C. L. Andersen, J. Bergstedt, J. H. von Stemann, M. Jørgensen, M.-H. E. Tang, M. Fontes, J. Bahlo, C. D. Herling, M. Hallek, J. Lundgren, C. R. MacPherson, J. Larsen and C. U. Niemann, "Machine learning can identify newly diagnosed patients with CLL at high risk of infection," *Nature Communications,* vol. 11, January 2020.

[9] J. D. Janizek, S. Celik and S.-I. Lee, "Explainable machine learning prediction of synergistic drug combinations for precision cancer medicine," May 2018.

[10] T. W. Campbell, M. P. Wilson, H. Roder, S. MaWhinney, R. W. Georgantas, L. K. Maguire, J. Roder and K. M. Erlandson, "Predicting prognosis in COVID-19 patients using machine learning and readily available clinical data," *International Journal of Medical Informatics,* vol. 155, p. 104594, November 2021.

[11] S. M. Lundberg, B. Nair, M. S. Vavilala, M. Horibe, M. J. Eisses, T. Adams, D. E. Liston, D. K.-W. Low, S.-F. Newman, J. Kim and S.-I. Lee, "Explainable machine-learning predictions for the prevention of hypoxaemia during surgery," *Nature Biomedical Engineering,* vol. 2, p. 749–760, October 2018.

[12] D. Wang, C. Zhang, B. Wang, B. Li, Q. Wang, D. Liu, H. Wang, Y. Zhou, L. Shi, F. Lan and Y. Wang, "Optimized CRISPR guide RNA design for two high-fidelity Cas9 variants by deep learning," *Nature Communications,* vol. 10, September 2019.

[13] K. Zhang, X. Liu, J. Shen, Z. Li, Y. Sang, X. Wu, Y. Zha, W. Liang, C. Wang, K. Wang, L. Ye, M. Gao, Z. Zhou, L. Li, J. Wang, Z. Yang, H. Cai, J. Xu, L. Yang, W. Cai, W. Xu, S. Wu, W. Zhang, S. Jiang, L. Zheng, X. Zhang, L. Wang, L. Lu, J. Li, H. Yin, W. Wang, O. Li, C. Zhang, L. Liang, T. Wu, R. Deng, K. Wei, Y. Zhou, T. Chen, J. Y.-N. Lau, M. Fok, J. He, T. Lin, W. Li and G. Wang, "Clinically Applicable AI System for Accurate Diagnosis, Quantitative Measurements, and Prognosis of COVID-19 Pneumonia Using Computed Tomography," *Cell,* vol. 181, p. 1423–1433.e11, June 2020.

[14] H. Ledford, "Millions of black people affected by racial bias in health-care algorithms," *Nature,* vol. 574, 2 October 2019.

[15] P. A. Noseworthy, Z. I. Attia, L. C. Brewer, S. N. Hayes, X. Yao, S. Kapa, P. A. Friedman and F. Lopez-Jimenez, "Assessing and Mitigating Bias in Medical Artificial Intelligence," *Circulation: Arrhythmia and Electrophysiology,* vol. 13, March 2020.

[16] Z. Obermeyer, B. Powers, C. Vogeli and S. Mullainathan, "Dissecting racial bias in an algorithm used to manage the health of populations," *Science,* vol. 366, p. 447–453, October 2019.

[17] N. Bhutta, A. Hizmo and D. Ringo, "How Much Does Racial Bias Affect Mortgage Lending? Evidence from Human and Algorithmic Credit Decisions," *SSRN Electronic Journal,* 2021.

[18] S. Lundberg and S.-I. Lee, "A Unified Approach to Interpreting Model Predictions," *31st Conference on Neural Information Processing Systems,* May 2017.

[19] D. Janzing, L. Minorics and P. Bloebaum, "Feature relevance quantification in explainable AI: A causal problem," in *<i>Proceedings of the Twenty Third International Conference on Artificial Intelligence and Statistics</i>, in <i>Proceedings of Machine Learning Research</i>*, 2020.

[20] M. Sundararajan and A. Najmi, "The Many Shapley Values for Model Explanation," in *<i>Proceedings of the 37th International Conference on Machine Learning</i>, in <i>Proceedings of Machine Learning Research</i>*, 2020.

[21] L. Merrick and A. Taly, "The Explanation Game: Explaining Machine Learning Models Using Shapley Values," September 2019.

[22] J. Pearl, "Causal diagrams for empirical research”, R-218-B," *Biometrika,* vol. 82, p. 669–710, 1975.

[23] C. Frye, D. de Mijolla, T. Begley, L. Cowton, M. Stanley and I. Feige, "Shapley explainability on the data manifold," in *International Conference on Learning Representations*, 2021.

[24] G. Hooker, L. Mentch and S. Zhou, "Unrestricted Permutation forces Extrapolation: Variable Importance Requires at least One More Model, or There Is No Free Variable Importance," May 2019.

[25] E. Štrumbelj, I. Kononenko and M. R. Šikonja, "Explaining instance classifications with interactions of subsets of feature values," *Data & Knowledge Engineering,* vol. 68, p. 886–904, October 2009.

[26] I. E. Kumar, S. Venkatasubramanian, C. Scheidegger and S. Friedler, "Problems with Shapley-value-based explanations as feature importance measures," February 2020.

[27] H. Chen, J. D. Janizek, S. Lundberg and S.-I. Lee, "True to the Model or True to the Data?," June 2020.

[28] J. Wang, J. Wiens and S. Lundberg, "Shapley Flow: A Graph-based Approach to Interpreting Model Predictions," October 2020.

[29] [Online]. Available: https://github.com/slundberg/shap. [Accessed 1 10 2021].

[30] T. Chen and C. Guestrin, "XGBoost: A Scalable Tree Boosting System," March 2016.

[31] W. Henry and Miller, *Plan and operation of the health and nutrition examination survey, United States, 1971-1973. DHEW publication no.(PHS)-Dept,* USA, 1973.

[32] F. Pedregosa, G. Varoquaux, A. Gramfort, V. Michel, B. Thirion, O. Grisel, M. Blondel, P. Prettenhofer, R. Weiss, V. Dubourg, J. Vanderplas, A. Passos, D. Cournapeau, M. Brucher, M. Perrot and E. Duchesnay, "Scikit-learn: Machine Learning in Python," *Journal of Machine Learning Research,* vol. 12, p. 2825–2830, 2011.

[33] [Online]. Available: https://github.com/Biodesix/EjectShapley. [Accessed 1 10 2021].

[34] S. Ghalebikesabi, L. Ter-Minassian, K. Diaz-Ordaz and C. Holmes, "On Locality of Local Explanation Models," June 2021.

[35] "Anaconda Documentation," *Anaconda Software Distribution,* 2020.

[36] J. D. Hunter, "Matplotlib: A 2D graphics environment," *Computing In Science & Engineering,* vol. 9, p. 90–95, 2007.

[37] M. L. Waskom, "seaborn: statistical data visualization," *Journal of Open Source Software,* vol. 6, p. 3021, 2021.

[38] H. Roder, C. Oliveira and L. Net, "Robust identification of molecular phenotypes using semi-supervised learning," *BMC Bioinformatics,* vol. 20, p. 273, 2019.

[39] D. Venet, J. Dumont and V. Detours, "Most random gene expression signatures are significantly associated with breast cancer outcome," *PLoS Comput Biol,* vol. 7, p. e1002240, 2011.

[40] W. Johnson, C. Li and A. Rabinovic, "Adjusting batch effects in microarray expression data using empirical Bayes methods," *Biostatistics,* vol. 8, p. 118–127, 2007.

# Supplement

Another set of synthetic data was generated with 13 features of varying expression difference (0, 0.25, 0.50, …, 3.0). A random classification forest was trained using 30 instances from each group and Shapley values (SVs) were calculated on an independent set, also with 30 instances in each group. We define SV~ as the product of the SVs and the classification of the instance ($\pm 1$). Hence, a positive SV~ corresponds to attribution in the same direction as classification. Means with associated error of the SV~ distributions were calculated across the entire set for each feature. This was done for a set with no correlations and for one with uniformly correlated features with off-diagonal covariances of 0.5. Figure E1 summarizes these results for TreeSHAP and Eject and also gives raw SV residuals comparing TreeSHAP to the Eject and Interventional approaches. We notice that for less informative features (lower expression difference), TreeSHAP assigns less attribution compared to Eject, with that trend reversing for the most informative features. Additionally, the Interventional approach gives much closer values to TreeSHAP than to Eject.

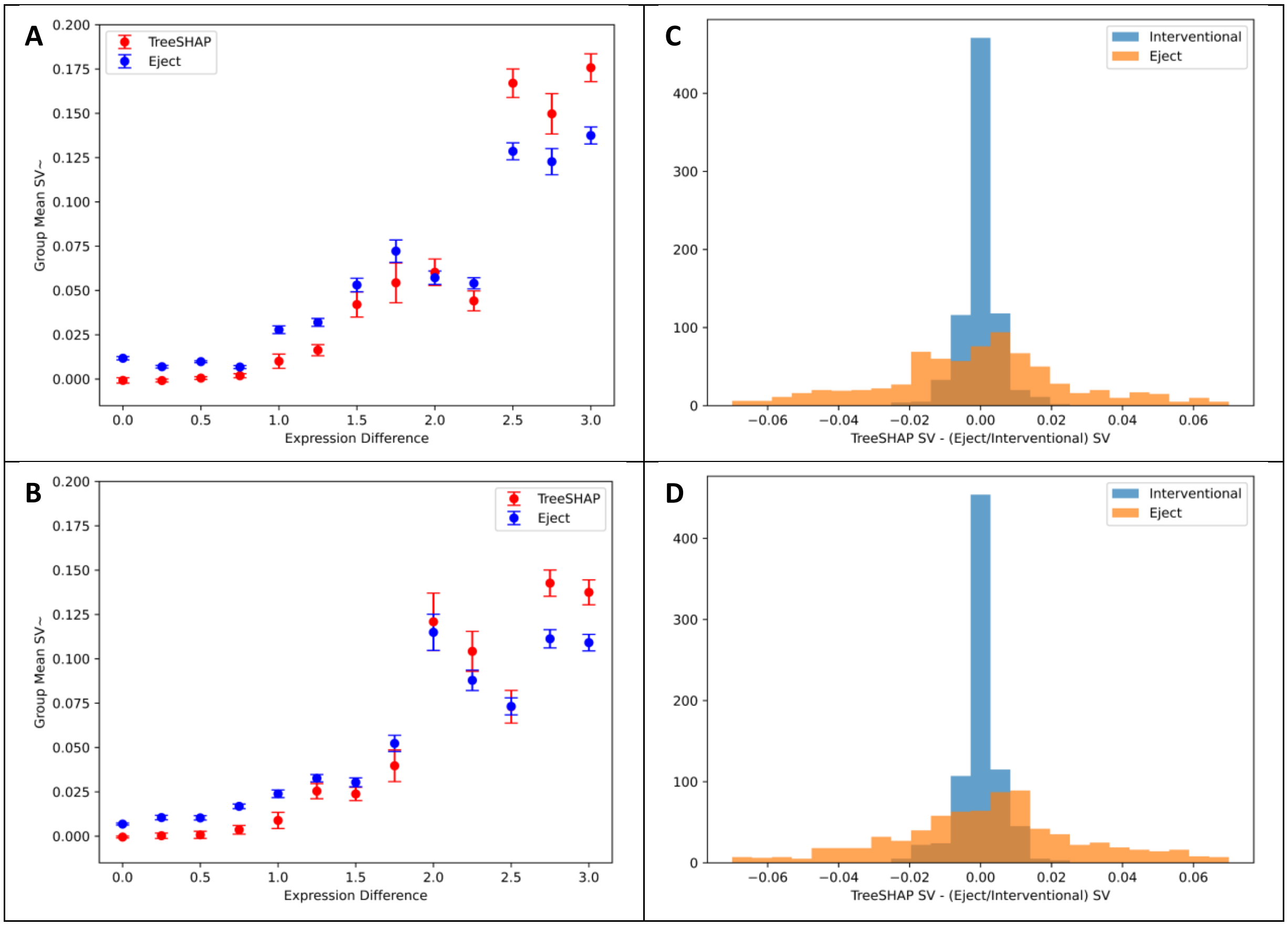

**Figure E1:** SV~ means and errors in the (**A**) uncorrelated and (**B**) correlated sets for TreeSHAP and Eject. SV residual comparing TreeSHAP to Interventional and Eject in the (**C**) uncorrelated and (**D**) correlated sets.

A SciKitLearn random classification forest with 1,000 trees, min_samples_leaf=5, and otherwise default parameters was trained to predict mortality (insert that 0 corresponds to prediction of death (or not) and 1 to the other ) on the NAHANESI I set using the same features and cohorts as was presented before. SVs were again calculated on the validation cohort, where accuracy was 83%. Figures E2 (A-C) give the distribution of these SVs for selected features. Figure E3 gives residuals in the SVs comparing TreeSHAP to Eject and Interventional. We see differences in both variance and shape of the distributions between Eject and TreeSHAP, with Interventional and TreeSHAP producing much more similar results than Eject.

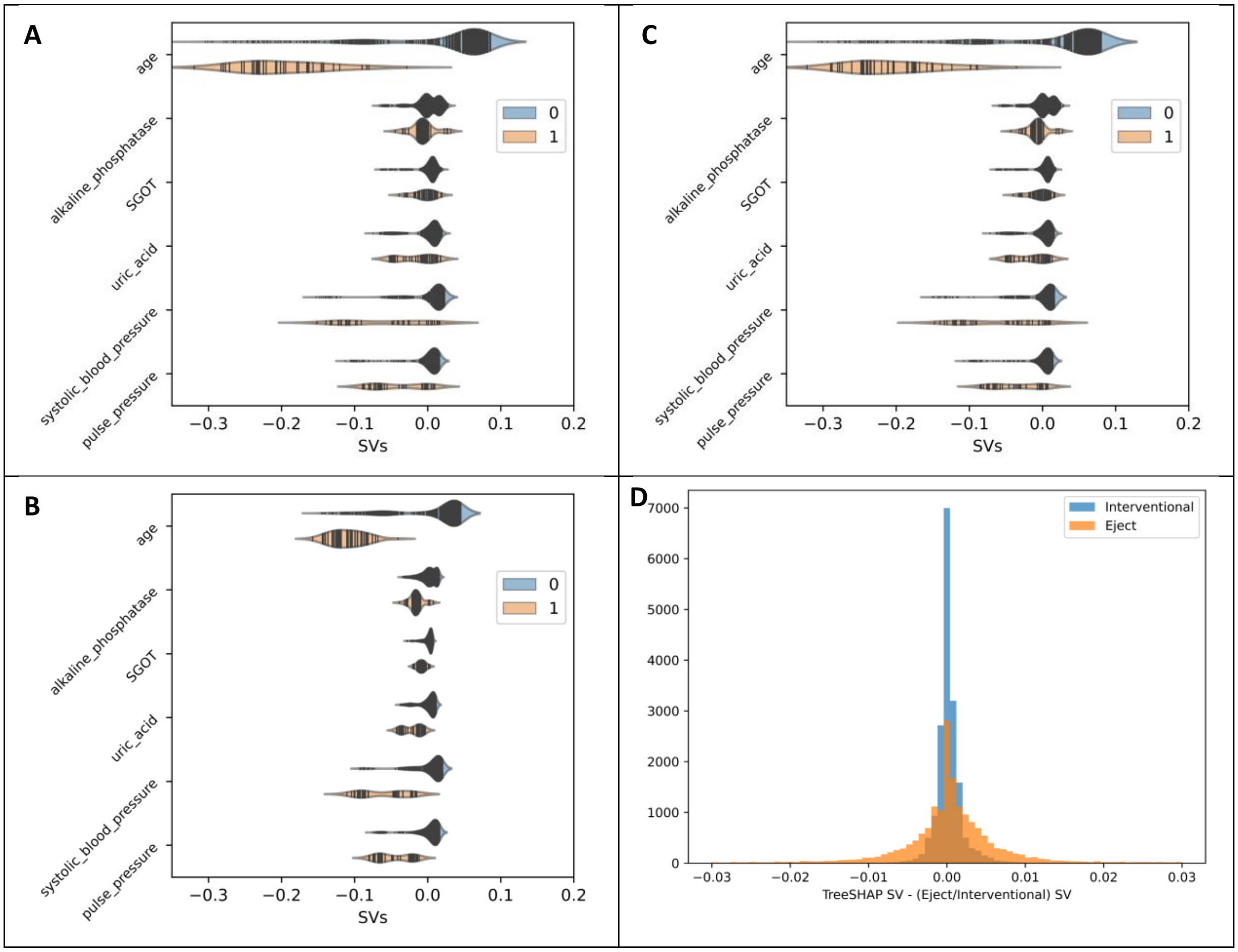

**Figure E2:** SVs for selected features using (**A**) TreeSHAP, (**B**) Eject, and (**C**) Interventional. Shaded areas represent a smoothed density distribution of the SVs with the vertical lines being each individual SV. SV residual comparing TreeSHAP to Interventional and Eject (**D**).